# A Hybrid Algorithm for Disparity Calculation From Sparse Disparity Estimates Based on Stereo Vision

Subhayan Mukherjee
Department of Information Technology
National Institute of Technology Karnataka
Surathkal, Mangalore, India
subhayan001@gmail.com

Ram Mohana Reddy Guddeti
Department of Information Technology
National Institute of Technology Karnataka
Surathkal, Mangalore, India
profgrmreddy@nitk.ac.in

*Abstract*—In this paper, we have proposed a novel method for stereo disparity estimation by combining the existing methods of block based and region based stereo matching. Our method can generate dense disparity maps from disparity measurements of only 18% pixels of either the left or the right image of a stereo image pair. It works by segmenting the lightness values of image pixels using a fast implementation of K-Means clustering. It then refines those segment boundaries by morphological filtering and connected components analysis, thus removing a lot of redundant boundary pixels. This is followed by determining the boundaries' disparities by the SAD cost function. Lastly, we reconstruct the entire disparity map of the scene from the boundaries' disparities through disparity propagation along the scan lines and disparity prediction of regions of uncertainty by considering disparities of the neighboring regions. Experimental results on the Middlebury stereo vision dataset demonstrate that the proposed method outperforms traditional disparity determination methods like SAD and NCC by up to 30% and achieves an improvement of 2.6% when compared to a recent approach based on absolute difference (AD) cost function for disparity calculations [1].

## I. INTRODUCTION

Disparity calculation based on stereo vision is an important problem in computer vision research [2] as it has applications in several areas of image processing in general, and particularly in robotic vision, 3D scene reconstruction, object detection and tracking, etc. The main challenge here is to generate accurate depth information of a scene by comparing the pixels of the left and right image of that scene. It is challenging, as individual pixels contain only colour and spatial information, and hence these represent low level image features [3]. Hence, to compare a stereo image pair effectively, there is a need for identifying appropriate high level image features and developing efficient methods to compare them with reasonable accuracy and speed.

Disparity of a pixel is inversely proportional to the distance of a point in the scene (represented by that pixel) from the camera. A disparity map or depth map represents the mapping of each pixel of an image to its corresponding disparity. To find the disparity of a pixel, we must first decide on a cost function to quantify the similarity between pixels of the left and right images of a stereo image pair. Then, we estimate its disparity by the relative displacement of the corresponding matching counterparts across the image pair. A crucial step in the disparity computation is finding matching pixels between the image pair, for which there are two basic approaches: block based and region based. Advantages and weaknesses of these existing solutions as analyzed in [1] are summarized in Table I.

TABLE I.     COMPARISON OF BASIC DEPTH ESTIMATION APPROACHES

|  | *Block-based Method* | *Region-based Method* |
|---|---|---|
| **Approach** | Depth estimation based on information contained in pixels and their surroundings. | Depth estimation based on optimal value of cost function for entire image regions of pixels having similar disparities. |
| **Strength** | High resolution depth maps. | Sharp edges on depth maps. |
| **Weakness** | Need to determine optimal block sizes for different images, or even different regions of the same image for creating accurate depth maps. | Unsuitable for finding gradually changing disparities, as all pixels constituting even a large region share a constant disparity. |

This motivates us to devise a method which can help us to overcome the limitations of these two basic approaches. Efforts have been made to improve them, as summarized in Table II.

TABLE II.     COMPARISON OF RECENT DEPTH ESTIMATION METHODS

| Work | Advantages | Limitations |
|---|---|---|
| *Choi [4]* | Bi-directional consistency check of disparity blocks can improve disparity estimates. | Comparison with similar algorithms using a standard dataset has not been done. |
| *Zhu and Yu [5]* | Self-adaptive block matching can increase the efficiency of searching, and reduce errors in matching and block-shape. | Presence of noise in the original left or right images can result in selection of incorrect size for block-matching (8x8 or 16x16). |
| *Wang and Zheng [6]* | Cooperative optimization can check or fix disparity errors. | Uses the time-consuming mean-shift algorithm for segmentation. |
| *Lu and Du [7]* | Harris corner point extraction and feature-matching yields better corresponding points than traditional methods. | Both left and right images are scanned for corner extraction and matching in the initial steps, making them time-consuming. |

To overcome the problems discussed above, in this paper we propose a method of disparity estimation by segmenting the lightness values of only left image pixels. It is done using a fast histogram-based implementation of K-Means clustering and by further refining segment boundaries using morphological filters and connected components analysis.

The core idea behind our method is to estimate disparities of only pixels lying on the refined segment boundaries. For this we use a block-based approach, taking large block sizes to get reliable disparity estimates in less number of computations. So, our method is scalable to high resolution stereo image pairs. Next, our disparity map reconstruction method described later estimates disparities of pixels lying inside segment boundaries.

Our method is different from [1] as they use the absolute difference (AD) cost function and we use SAD, and moreover, they use a one-dimensional colour-based segmentation process of individual rows of pixels, whereas we use two-dimensional intensity based segmentation of the left image. Our method is distinct from [3] as they use object-based segmentation using colour, spatial and shape information, but we use only the 'L' values of pixels, making the process faster.

Two key contributions of the proposed work are as follows:
- To the best of our knowledge, we are the first to propose a method for estimation of stereo disparity by segmentation of only lightness values of left image pixels, unlike other methods that use colour, texture and shape characteristics.
- To the best of our knowledge, this is the first paper which employs morphological filters and connected component analysis to obtain sparse, but accurate disparity estimates.

The rest of this paper is organized as follows: Section II describes our proposed algorithm; Section III discusses the step wise outputs of our algorithm and compares its generated depth map with that of two established methods, NCC and SAD and a recent approach [1] using ground-truth depth maps from the Middlebury stereo vision dataset [8]. Section IV concludes our paper with possible future directions of the proposed work.

## II. PROPOSED WORK

Our proposed algorithm calculates the depth map following the flowchart in Fig. 1. The steps are explained subsequently.

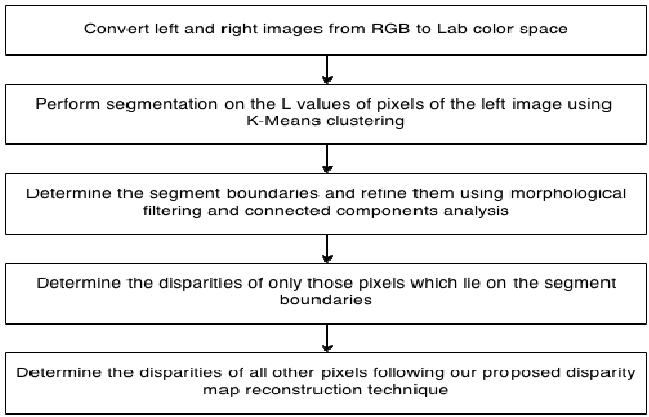

Fig. 1. Flowchart of Our Proposed Algorithm

### A. Color Space Conversion

Most imaging equipment store captured images in the RGB format which models the output of physical devices, but does not properly approximate human vision. The Lab colour space was developed to aspire for uniformity in human perception. Human eyes are more sensitive to changes in brightness than colour. The L component of the Lab colour space closely matches the human perception of lightness. Hence, like many other image processing algorithms, by applying the pertinent transformations discussed in [9], we convert the left and right images from RGB to the Lab colour space and retain only the L values of its pixels for further processing.

### B. Segmentation

We perform segmentation on the L values of the left image pixels using a fast implementation of the K-Means clustering algorithm. Since we are using a one-dimensional feature vector (containing only the pixel's L value) to represent the image pixels for clustering, we build a histogram of the L values and use that histogram instead of the actual pixel values for the subsequent iterations of K-Means clustering. Thus, the runtime of each iteration is significantly reduced, as the histogram has a small fixed number of bins, and we perform clustering on those bins instead of the actual pixels. Since there exists a one-to-one correspondence between each pixel and the bin to which it has been mapped, at the end of the clustering process, we are easily able to identify the cluster to which the pixel has been assigned by the cluster to which its associated bin has been assigned.

### C. Segment Boundary Detection and Refinement

Segment boundary detection is achieved by comparing the cluster assignment of each pixel with that of its 8-connected pixels (i.e. its Moore neighbourhood). If either of them is found to be different, we mark the pixel as '1' (belonging to a segment boundary), else we mark it as '0' (not belonging to any segment boundary). Thus, this step generates the "boundary map" from the segmented left image.

However, the above approach also falsely identifies many pixels as belonging to segment boundaries due to limitations imposed by clustering accuracy, as the clustering is done only based on the pixels' lightness (L) values and not on their colour components (a, b) or their spatial locations (x, y) in the image. So, we apply two morphological filters to refine the boundary map by removing such noisy pixels, in the following order:

*Fill*: Fills isolated interior pixels (individual 0's that are surrounded by 1's), such as the centre pixel in this pattern:

$$\begin{matrix} 1 & 1 & 1 \\ 1 & 0 & 1 \\ 1 & 1 & 1 \end{matrix}$$

*Remove*: Removes interior pixels, i.e., sets a pixel to '0' if all its 4-connected neighbours are '1', thus leaving only the boundary pixels on.

Further, we use connected components analysis and remove small artefacts in the boundary map due to segmentation errors, by creating a frequency distribution of connected components present in the boundary map, ordered by the number of pixels constituting each connected component. Lastly, we remove the smallest connected components which contribute about 4% of the total number of pixels present in the boundary map.

### D. Disparity Measurement of Boundary Pixels

We use the SAD (Sum of Absolute Differences) cost function to determine only the disparities of boundary pixels, using the L values of the left image pixels (target) and the L values of the right image pixels (reference). Thus, from such sparse disparity measurements, we compute a partial disparity map, from which we will create the full disparity map using our a disparity map reconstruction technique explained next. It should also be noted here, that by "boundary" pixels, we are also (implicitly) referring to the pixels of the left image that map to the left and right borders of the disparity map.

### E. Disparity Map Reconstruction from Boundaries

Our disparity map reconstruction algorithm scans through each row of the partially computed disparity map and computes the remaining disparities based on disparities that have already been calculated. Our algorithm operates in two stages:

#### 1) Disparity Propagation ('Fill' Stage):

In the first stage, we scan the disparity map row-wise, left to right—whenever we consecutively encounter two boundary pixels with identical disparity values, we 'fill' the intermediate pixels of that row falling between the segment boundaries with that disparity value. This reflects our assumption – that the pair of points in consideration actually belong to the same object in the original image. So, all the intermediate pixels of the image in that row falling between those points also belong to that same object, and hence, should have similar disparity values. However, we make an exception while 'filling' the disparity values of pixels near the left and right end of each row—we continue 'filling' the left and right ends of each row with the disparity values of the nearest border pixel till we encounter a boundary pixel. The process is explained using the pseudo-code below; $disp\_map$ is the matrix containing the disparities of boundary pixels and '-1' for all non-boundary pixels, $n\_rows$ and $n\_cols$ are the number of rows and columns in $disp\_map$.

**Algorithm 1** Disparity Propagation along Scan Lines

```
1:     for i = 1 → n_rows do
2:         prev_disp ← -1
3:         prev_indx ← 2
4:         for j = 2 → n_cols − 1 do
5:             if disp_map(i, j) > -1 then
6:                 if prev_disp > -1 then
7:                     if disp_map(i, j) == prev_disp then
8:                         for k = prev_indx → j − 1 do
9:                             disp_map(i, k) ← prev_disp
10:                        end
11:                    end
12:                else
13:                    border_disp ← disp_map(i, 1)
14:                    for k = prev_indx → j − 1 do
15:                        disp_map(i, k) ← border_disp
16:                    end
17:                end
18:                prev_disp ← disp_map(i, j)
19:                prev_indx ← j + 1
20:            end
21:        end
22:        border_disp ← disp_map(i, n_cols)
23:        for k = prev_indx → n_cols − 1 do
24:            disp_map(i, k) ← border_disp
25:        end
26:    end
```

#### 2) Estimation from Known Disparities ('Peek' Stage):

In the second stage, for all pixels whose disparities have not yet been determined, we estimate their disparities by 'peek'-ing at the disparity values of their neighbouring pixels.

**Algorithm 2** Disparity Estimation from Known Disparities

```
1:     for each row of disp_map starting from its top do
2:         for each cell of this row, starting from its left do
3:             if disparity of this cell has not been already determined, then
4:                 disp_n ← {known disparities in its neighbouring cells}
5:                 disparity of this cell ← statistical mode of disp_n
6:             end
7:         end
8:     end
```

### III. RESULTS AND DISCUSSION

To evaluate the stereo depth estimation performance of our approach vis-a-vis those of some existing algorithms, we chose three pairs (left and right) of images from the Middlebury data-set, viz. Tsukuba, Sawtooth and Venus and ran our algorithm on those three image pairs. We then compared the depth maps generated by our algorithm against the corresponding ground truth depth maps provided in the data-set, as well as with the depth maps generated by two established methods (SAD and NCC) and a recent one [1] on those same three image pairs.

We present the results of the said comparisons below, along with the outputs of each step of our proposed approach for the Tsukuba image pair, to demonstrate how our algorithm works.

### A. Discussion of Step-wise Outputs of Proposed Approach

For demonstrating the internal workings of our algorithm, in this sub-section, we consider the Tsukuba image pair from the Middlebury standard dataset and discuss the intermediate outputs after each step as described in the previous section. We present the values used for the three parameters in Table III.

TABLE III. PARAMETER VALUES FOR THE THREE IMAGE PAIRS

| Parameter | Tsukuba | Sawtooth | Venus |
|---|---|---|---|
| Number of clusters (K) for K-Means clustering | 10 | 12 | 10 |
| Size of block size for cost aggregation (odd) | 7 x 7 pixels | 9 x 9 pixels | 9 x 9 pixels |
| Size of window for disparity estimation of uncertain regions (odd) | 5 x 5 pixels | 5 x 5 pixels | 5 x 5 pixels |

#### 1) Colour Space Conversion

*Process:* Left and right input images are converted from the RGB to the CIE-Lab color space.

*Results:* Each pixel of both images has a L value for its lightness component. Fig. 2 (left) shows the converted left input image.

#### 2) Segmentation

*Process:* Converted left input image image is fed to the fast implementation of K-Means clustering algorithm as discussed earlier.

*Results:* The result is shown in Fig. 2 (right).

*Observations:* Small variations in 'L' values (noise) have been abosorbed into single coherent segments representing image regions of pixels belonging to the same object, with similar values of disparity. However, since we have segmented the left image based on just the pixels' 'L' values, ignoring their spatial locations (to reduce the feature vector size), we must refine the boundaries of objects defined by the clusters above, so that the spatial distribution of the 'L' values are also taken into account.

*3) Segment Boundary Detection and Refinement*

*Process:* We detect and refine the segment boundaries using the approach desribed in the previous section.

*Results:* The outputs are shown in Fig. 3, with output of segment boundary point detection, followed by that of morphological filtering, and lastly, output of connected components analysis at the bottom.

*Observations:* We can infer from Fig. 3b, how the redundancies in segment boundary detection have been removed from Fig. 3a, and from Fig. 3c, how the small artefacts in Fig. 3b have been removed by the connected components analysis technique.

For the Tsukuba image pair, the segment boundary refinement reduces the number of boundary pixels by nearly 53% for the parameter values of Table III, such that only 19% of the number of pixels of the left image are used for disparity calculations. This greatly reduces the number of disparity computations in the next step.

*4) Disparity Measurement of Boundary Pixels*

*Process:* Disparities of the boundary pixels are determined by using the SAD algorithm.

*Results:* The result in shown in Fig. 4 (left).

*Observations:* Boundaries of objects closeer to the camera (like the lamp and the head of the statue) are having a higher intensity (greater value of stereo disparity). This is in direct agreement with the reality that value of disparity of a pixel is inversely proportional to its distance (from the camera lens / human eyes).

*5) Disparity Map Reconstruction from Boundaries*

*Process:* Disparities of the segmented regions in the image are determined from the disparities of their boundaries using the two-stage disparity map reconstruction method discussed earlier.

*Results:* The results are shown in Fig. 4 (right).

### B. Comparison with Ground Truth Disparity Map

For evaluating the disparity calculations performed by our proposed method, we consider the approach of computing error statistics with respect to the ground truth image available with the Middlebury dataset, and we choose the quality metric as the percentage of bad matching pixels (B) given by Eq. 1,

$$B = \frac{1}{N} \sum_{(x,y)} (|d_C(x,y) - d_T(x,y)| > \delta_d) \quad (1)$$

where $d_C(x, y)$ refers to the computed disparities and $d_T(x, y)$ to the ground truth disparities, and $\delta_d$ denotes the disparity error tolerance, which is taken as '1.0' as per published works [8].

Fig. 5 presents a comparison of the disparity map generated by our method (left) with Middlebury's ground truth disparity map (right). The black regions in disparity maps denote regions for which disparities have not been compared. Moreover, the percentage of bad matching pixels was calculated as 11.33%.

### C. Comparison with other Disparity Computation Methods

Table IV presents a performance comparison of our disparity calculation method with two established methods, viz. Sum of absolute differences (SAD) and normalized cross-correlation (NCC), both of which have been used in numerous traditional algorithms for stereo disparity calculations, and a recent work [1] which uses the absolute difference (AD) cost function to predict disparities. We use the same quality mertic, percentage of bad matching pixels, for quantitative comparison of depth maps generated by our method, SAD, NCC and [1].

The results clearly demonstrate that, our proposed method performs better than the traditional NCC and SAD methods by a large margin, and outperforms even recent methods like [1].

Further, the percentage of left image pixels used for stereo disparity calculations was about 19% for Tsukuba, around 18% for Sawtooth and 16% for Venus image pairs, implying that our algorithm is scalable to high resolution stereo image pairs.

Fig. 6 shows disparity maps generated by our method (left) and Middlebury's ground truth disparities (right) for Sawtooth and Venus. All other disparity maps supporting performance comparison data presented in Table IV can be found in [1].

TABLE IV. PERFORMANCE COMPARISON OF PROPOSED ALGORITHM

|  | *Tsukuba %* | *Sawtooth %* | *Venus %* |
|---|---|---|---|
| NCC | 41.4 | 9.92 | 17.4 |
| SAD | 36.9 | 11.9 | 24.5 |
| Zhen Zhang et al. [1] | 13.9 | 7.22 | 6.12 |
| Proposed Algorithm | 11.3 | 6.22 | 5.71 |

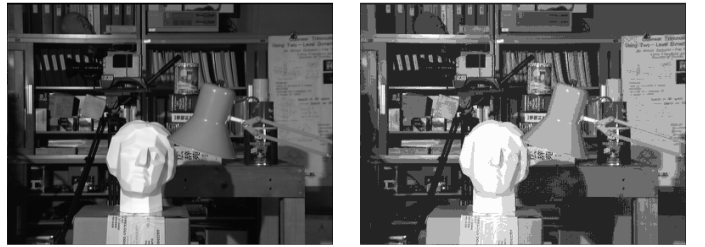

Fig. 2. Segmentation of 'L' Values of Left Image using K-Means

a) 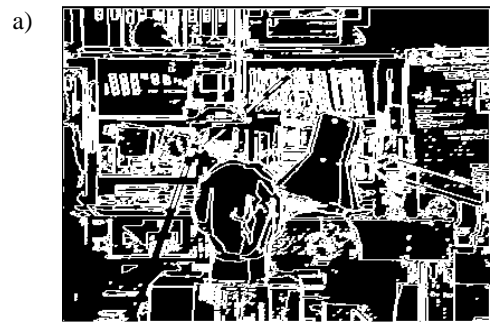

b)

c)

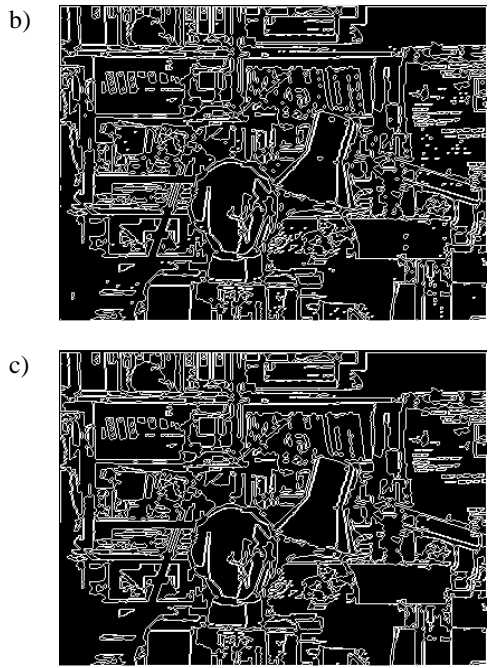

Fig. 3. Segment Boundary Detection and Refinement

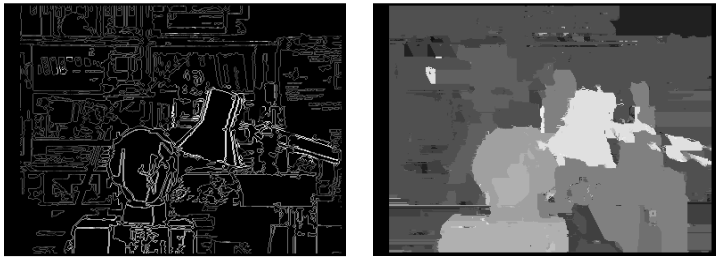

Fig. 4. Disparity Map Reconstruction from Boundary Disparities

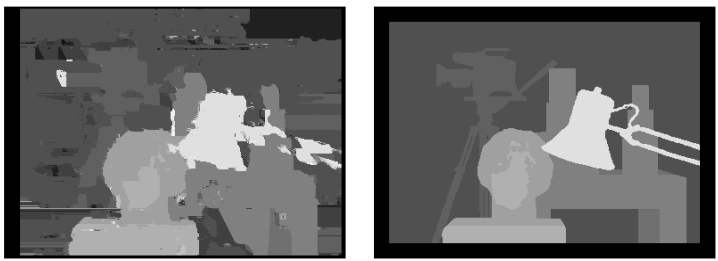

Fig. 5. Resultant Disparity Map as Compared to Middlebury's Ground Truth

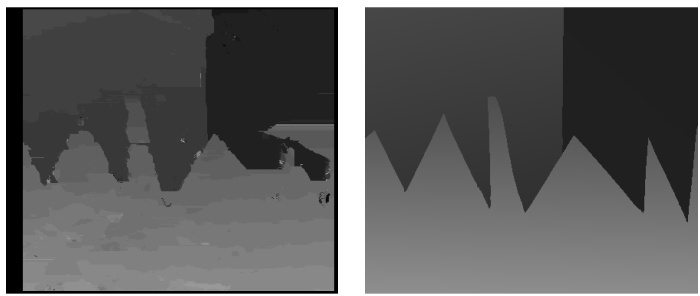

Fig. 6. Sawtooth (top) and Venus (bottom) Disparity Maps for Proposed Algorithm (left) and Middlebury's Ground Truth Disparity Maps (right)

## IV. CONCLUSION AND FUTURE WORK

In this paper, we have proposed an approach for disparity estimation based on segmentation of the lightness values of the image pixels and subsequent refinement of segment boundaries by morphological filtering and connected components analysis, followed by determination of the boundaries' disparities using the SAD cost function, and lastly, disparity map reconstruction from the boundaries' disparities using our adopted approach.

Future work will be focusing mainly on reducing incorrect disparity estimates of boundary pixels and on improving the disparity map reconstruction technique.

REFERENCES

[1] Zhen Zhang; Yifei Wang; Dahnoun, N., "A novel algorithm for disparity calculation based on stereo vision," Education and Research Conference (EDERC), 2010 4th European , vol., no., pp.180,184, 1-2 Dec. 2010

[2] N. Lazaros, G. Sirakoulis, and A. Gasteratos. Review of Stereo Vision Algorithms: From Software to Hardware. International Journal of Optomechatronics, 2(4):435–462, 2008.

[3] Jun Xiao; Linyuan Xia; Liqun Lin; Zhentao Zhang, "A segment-based stereo matching method with ground control points," Environmental Science and Information Application Technology (ESIAT), 2010 International Conference on , vol.3, no., pp.306,309, 17-18 July 2010
doi: 10.1109/ESIAT.2010.5568363

[4] Kang-Sun Choi, "Hierarchical block-based disparity estimation," Consumer Electronics (GCCE), 2012 IEEE 1st Global Conference on , vol., no., pp.493,494, 2-5 Oct. 2012, doi: 10.1109/GCCE.2012.6379668

[5] Shiping Zhu; Yang Yu, "Virtual View Rendering Based on Self-adaptive Block Matching Disparity Estimation," Industrial Control and Electronics Engineering (ICICEE), 2012 International Conference on , vol., no., pp.947,950, 23-25 Aug. 2012, doi: 10.1109/ICICEE.2012.251

[6] Zeng-Fu Wang; Zhi-Gang Zheng, "A region based stereo matching algorithm using cooperative optimization," Computer Vision and Pattern Recognition, 2008. CVPR 2008. IEEE Conference on , vol., no., pp.1,8, 23-28 June 2008, doi: 10.1109/CVPR.2008.4587456

[7] Di Lu; Yu Du, "A two-step stereo correspondence algorithm based on combination of feature-matching and region-matching," Strategic Technology (IFOST), 2013 8th International Forum on , vol.2, no., pp.51,55, June 28 2013-July 1 2013, doi: 10.1109/IFOST.2013.6616858

[8] Scharstein, D.; Szeliski, R.; Zabih, R., "A taxonomy and evaluation of dense two-frame stereo correspondence algorithms," Stereo and Multi-Baseline Vision, 2001. (SMBV 2001). Proceedings. IEEE Workshop on , vol., no., pp.131,140, 2001, doi: 10.1109/SMBV.2001.988771

[9] Tkalcic, M.; Tasic, J.F., "Colour spaces: perceptual, historical and applicational background," EUROCON 2003. Computer as a Tool. The IEEE Region 8 , vol.1, no., pp.304,308 vol.1, 22-24 Sept. 2003
doi: 10.1109/EURCON.2003.1248032